\documentclass[10pt,journal,compsoc]{IEEEtran}

\ifCLASSOPTIONcompsoc
  \usepackage[nocompress]{cite}
  \usepackage{cite}
  \usepackage{amsmath,amssymb,amsfonts}
  \usepackage{graphicx}
  \usepackage{textcomp}
  \usepackage{xcolor}
  \usepackage{array}
  \usepackage{multirow}
  \usepackage {subcaption}
  \usepackage{booktabs}
  \usepackage{algorithm}
  \usepackage{algorithmic}
\else
  \usepackage{cite}
\fi

\ifCLASSINFOpdf
 
\else
 
\fi

\begin{document}

\title{Deep Visual Waterline Detection within Inland Marine Environment}

\author{Jing Huang, Hengfeng Miao, Lin Li, Yuanqiao Wen, Changshi Xiao
\IEEEcompsocitemizethanks{\IEEEcompsocthanksitem Jing Huang is with the School of Computer Science and Technology, Wuhan University of Technology, Wuhan, China, 430063.\protect\\
E-mail: huangjing@whut.edu.cn
\IEEEcompsocthanksitem Hengfeng Miao, Lin Li are with the School of Computer Science and Technology, Wuhan University of Technology. Yuanqiao Wen, Changshi Xiao are with the School of Navigation, Wuhan University of Technology.}

\thanks{}}

\markboth{}%
{Shell \MakeLowercase{\textit{et al.}}: Bare Demo of IEEEtran.cls for Computer Society Journals}

\IEEEtitleabstractindextext{%
\begin{abstract}
Waterline usually plays as an important visual cue for maritime applications. However, the visual complexity of inland waterline presents a significant challenge for the development of highly efficient computer vision algorithms tailored for waterline detection in a complicated inland water environment. This paper attempts to find a solution to guarantee the effectiveness of waterline detection for inland maritime applications with general digital camera sensor. To this end, a general deep-learning-based paradigm applicable in variable inland waters, named DeepWL, is proposed, which concerns the efficiency of waterline detection simultaneously. Specifically, there are two novel deep network models, named WLdetectNet and WLgenerateNet respectively, cooperating in the paradigm that afford a continuous waterline image-map estimation from a single captured video stream. Experimental results demonstrate the effectiveness and superiority of the proposed approach via qualitative and quantitative assessment on the concerned performances. Moreover, due to its own generality, the proposed approach has the potential to be applied to the waterline detection tasks of other water areas such as coastal waters. 
\end{abstract}

}

\maketitle

\IEEEdisplaynontitleabstractindextext

\IEEEpeerreviewmaketitle

\IEEEraisesectionheading{\section{Introduction}\label{sec:introduction}}
With the development of computer vision techniques, many waterline detection approaches by virtue of general digital camera have been proposed. Most of them aim at the detection of coastal waterline in sea areas, e.g. [1-4], and there are also a few work focusing on inland waterline, e.g. [5]. When applied in inland waters, the detection effects of these approaches tend to be vulnerable to the variations of environmental factors (e.g., weather conditions like fog, snow or rain, illumination conditions like shadow, reflection or water glint, the shapes of waterlines, as well as the viewpoints of cameras). The reason is that the erratic environmental factors usually engender more visual complexity on inland waterline. For example, the visual information of the background surrounding inland waterline might become more confusing due to the change of illumination. Correspondingly, the stability of existing approaches is prone to be disturbed in such changeable and complicated inland water scenarios. 

Existing vision-based waterline detection approaches generally share a pipeline that consists of two relevant processes, i.e., waterline-relevant feature representation and discriminative strategy (or algorithm) for final identification respectively. Nevertheless, current proposals for the two relevant processes in these vision-based approaches tend to overwhelm the stability of the approaches themselves, due to their deficiencies in the robustness against variable inland water environments: {\romannumeral 1}) The current proposals for representing waterline-relevant features are hand-crafted that largely depend on specific prior knowledge or statistical assumptions, e.g., [2,3,6,7], whereas the applied prior knowledge or assumptions cannot hold in all cases; {\romannumeral 2}) The currently used discriminative strategies (or algorithms) to finalize waterline identification mostly work in particular water conditions, which have little consideration for coping with the visual versatility of waterline caused by the variations of environmental factors, such as [5,8].

\begin{figure}[htbp]
\centerline{\includegraphics[width=9cm]{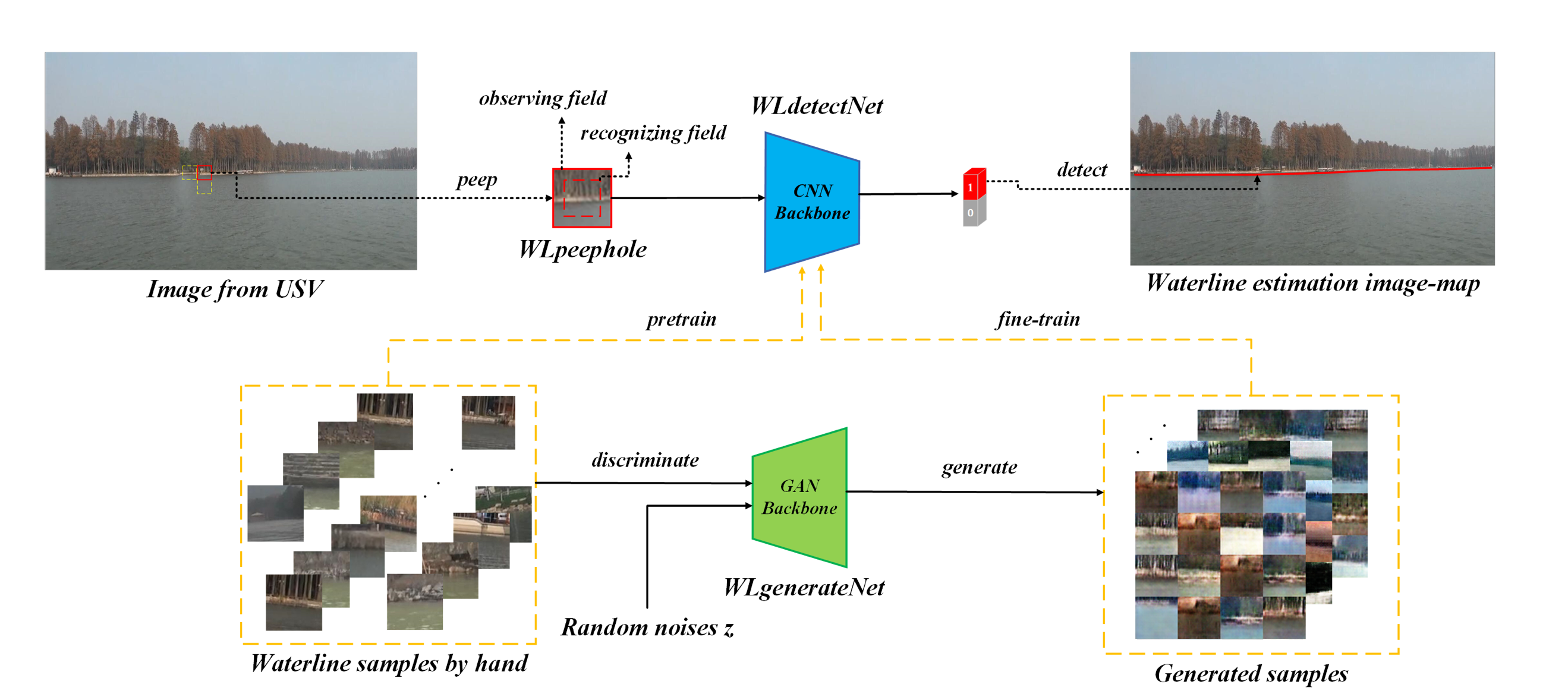}}
\caption{The diagram of DeepWL for waterline detection.}
\label{fig1}
\end{figure}

To meet the challenging task, this paper aims to guarantee the effectiveness of waterline detection for inland maritime applications, such as unmanned surface vehicles (USVs) mounted with camera sensor patrolling in variable inland water environments. Specifically, we make an attempt to improve the robustness and stability of waterline detection for diverse cases by proposing a general deep-learning-based paradigm for inland marine applications, named DeepWL, which concerns the efficiency of waterline detection simultaneously. As illustrated in Fig.\ref{fig1}, the proposed paradigm consists of two cooperative deep neural network models. One, termed WLdetectNet, is customized as the primary network (i.e., a learning model) for our deep visual waterline detection by exploiting convolutional neural networks (CNNs [9]). The other one, named WLgenerateNet, is built upon generative adversarial networks (GANs) [10] that serves as the auxiliary network (i.e., another learning model) of WLdetectNet. 

As the mainstay of this paradigm, the WLdetectNet is modeled as an end-to-end deep convolutional neural network to directly achieve the identification of candidate waterlines in each captured image, rather than separating waterline-relevant feature representation from its related discriminative strategy (or algorithm), and without resorting to image preprocessing. To improve the accuracy of waterline detection, we innovatively devise two significant schemes for the materialization of WLdetectNet by following the same architectural principles with many modern deep CNNs (e.g., [11], [12]), which are actually dedicated to the improvement of the representational capability of this deep learning model for varied waterlines. At the same time, owing to its architectural characteristics benefiting from the two specialized schemes, WLdetectNet lays stress on the efficiency of waterline detection as well. What is more, to improve the robustness of WLdetectNet for variable inland water environments, another deep neural network, i.e., WLgenerateNet, is built to assist the improvement of the generalization ability of WLdetectNet in diverse cases by exploiting the classical generative adversarial networks, such as  [13] and [14]. In brief, the overall design for the proposed paradigm is inspired by the current great successes of deep learning techniques in computer vision applications.

In addition, we define relevant metrics specialized for the quantitative evaluation of visual waterline detection approach on related performances, and conduct empirical investigations on the effectiveness and superiority of the proposed approach via qualitative and quantitative assessment. Compared with other alternative approaches, the proposed approach achieves better robustness and stability in the presence of environmental noises in variable inland water. 

\section{Methodology}
\label{sec2}
As illustrated in Fig.\ref{fig1}, the paradigm comprises two collaborative deep neural networks, i.e., WLdetectNet and WLgenerateNet. Thus, in this section, we first highlight the architectural details of the two significant deep networks, and then describe their training methods. Finally, based on the proposed paradigm, we present an algorithm to achieve waterline detection. 

\subsection{The main network WLdetectNet}
In this paradigm, we specify the vision-based waterline detection as an end-to-end binary classification model, which integrates the waterline-relevant feature representation and its subsequent waterline discriminator in a customized deep convolutional neural network, i.e., WLdetectNet. To improve the accuracy of waterline detection, we deliberately devise the following two specialized schemes to construct the deep model. 

\emph{1) Building a perceptive block as the receptive field of WLdetectNet}:
Given the visual complexity of waterline in varied inland water environments, we intentionally build a block to specialize in perceiving the contextual information relevant to waterline segments, and make use of the perceptive block as the receptive field of WLdetectNet, i.e., as the first layer of the deep main network in our paradigm DeepWL. As shown in Fig.\ref{fig1}, the premeditated block, termed WLpeephole, is designated to be an image region of size $r \times r$, which consists of two different size fields (i.e., $r\times r$ and $s\times s$, besides $r>s$). Specifically, in order to more conveniently and precisely distinguish various segments relevant to waterline, i.e., waterline segments, we take advantage of two different scale squares with a same central point, called \emph{observing field} ($r\times r$) and \emph{recognizing field} ($s\times s$) respectively, and further mandate the candidate segments of waterline to emerge only in the smaller square (i.e., \emph{recognizing field}). Correspondingly, the area between \emph{observing field} and \emph{recognizing field}, namely the area \emph{observing field} surrounding \emph{recognizing field}, may be filled with various contextual information associated with a waterline segment, e.g., water streak, plants or buildings in waterfront. 

Intuitively, the special design on the block can draw visual attention to the waterline within a receptive field. In practice, by feeding an image patch in accordance with the block-based design into our deep network WLdetectNet, we can get hold of the more discriminative waterline feature that avails final accurate decision-making on whether this image patch contains waterline or not. The reason is that the block based on the design carries better characteristics for making distinctions between waterline and non-waterline by paying attention to necessary contexts associated with waterline in such a receptive field. And the disparity information mingled in a block facilitates WLdetectNet to bolster the ability of the deep network regarding waterline-relevant feature representation. Besides, in our waterline detection algorithm (Algorithm 1), the perceptive block WLpeephole actually behaves as a peephole that successively diagnoses each region across an image captured by USVs to tell whether the current diagnosed region (i.e., current receptive field of WLdetectNet) contains waterline or not.

\emph{2) Deepening WLdetectNet to improve its own representational capability}:
It is generally believed that improving their own representational capability of learning models is a predominant means to bolster the specific performance of tasks based on machine learning, such as accuracy on prediction [15,16]. Inspired by the great success of extending the depth of CNNs, e.g., [17,18,19,20], in this paper, we innovatively constitute a rather deep architecture for WLdetectNet to guarantee its robust representational capability, thus improving the accuracy of the main network for waterline detection. A complete description of its architectural specification is presented in Table \ref{tab1}. It is worth mentioning that, similar to many modern variants of CNNs, some critical architectural principles are adopted in the construction of the deep network. For example, to extend the depth of WLdetectNet, we repeatedly exploit several structural modules in the residual branch of WLdetectNet like ResNet [17]. Then, to facilitate training such a deep network, we similarly take advantage of the shortcut path to back-propagate gradients. Meanwhile, in order to alleviate the information loss in such a deep network as much as possible to ensure the effectiveness and efficiency of this deep network in diverse cases, within each of the repeated structural modules, we attempt to successively make use of pointwise group convolution (i.e., PGconv), channel shuffle operation (i.e., Shuffle), depthwise convolution (i.e., Dwconv), point convolution (i.e., Pconv), global average pooling (i.e., GAP), fully connected operations (i.e., FC, with a Relu and Sigmoid activation, respectively) and channel-wise scaling operation (i.e., Scale) to enrich and equalize the information flow in the main network of our proposed paradigm. Especially, the reasonable utilization of PGconv and Dwconv in our architectural design of WLdetectNet benefits reducing the number of network parameters and computational complexity, which are crucial for guaranteeing the efficiency of the deep network. Actually, despite being deepened, WLdetectNet has little extra computational cost, about 3.12 MFLOPs that is very suitable for computationally limited applications. In addition, the channel operation Shuffle adopted in our architectural design also contributes to ensure the robust representational capability of WLdetectNet by equalizing information flow in such a deep network.

\begin{table}[htbp]
\scriptsize
\caption{The deep architecture of WLdetectNet}
\begin{center}
\begin{tabular}{p{50pt}<{\centering}p{33pt}<{\centering}p{28pt}<{\centering}p{95pt}}
\toprule[2pt]
\textbf{Layers}&\textbf{Output Size}&\textbf{Repeated}&\multicolumn{1}{c}{\textbf{Operations}} \\
\toprule[2pt]

An image (captured by USVs)&3x64x64& &{Sampling based on WLpeephole (i.e., by 64x64) as the first layer} \\

\hline

Input layer&64x64x64&1&\tabincell{l}{3x3, 64conv, stride 1} \\

\hline

Module-1&64x64x64&8&\tabincell{l}{1x1, 32PGconv, stride 1, group 4\\
					Shuffle, group 4\\
					3x3, 32Dwconv, stride 1\\
					1x1, 64Pconv, stride 1, group 4\\
					GAP, FC, FC} \\

\hline
Module-2&128x64x64&1&\tabincell{l}{1x1, 64PGconv, stride 1, group 4\\
													Shuffle, group 4\\
													3x3, 64Dwconv, stride 1    \\       
													1x1, 128Pconv, stride 1, group 4\\
													GAP, FC, FC\\
													1x1, 128conv, stride 1 (shortcut\\ path)} \\
\hline
Module-3&128x64x64&3&\tabincell{l}{1x1, 64PGconv, stride 1, group 4\\ 
													Shuffle, group 4\\
													3x3, 64Dwconv, stride 1     \\     
													1x1, 128Pconv, stride 1, group 4\\
													GAP, FC, FC} \\
\hline
Module-4&256x64x64&1&\tabincell{l}{1x1, 128PGconv, stride 1, group 4\\
													Shuffle, group 4\\
													3x3, 128Dwconv, stride 1         \\
													1x1, 256Pconv, stride 1, group 4\\
													GAP, FC, FC\\
													1x1, 256conv, stride 1 (shortcut\\ path)} \\

\hline

Module-5&256x64x64&3&\tabincell{l}{1x1, 128PGconv, stride 1, group 4\\
													Shuffle, group 4\\
													3x3, 128Dwconv, stride 1         \\
													1x1, 256Pconv, stride 1, group 4\\
													GAP, FC, FC} \\

\hline

\multirow{2}{*}{Output layer} 
& {2x1x1} & {} & {64x64, 2Convolution,stride 1}   \\

\cline{2-4}

& {1D} & {} & \multicolumn{1}{c}{Softmax} \\
\toprule[2pt]
\end{tabular}
\label{tab1}
\end{center}
\end{table}

As shown in Table \ref{tab1}, WLdetectNet uses individual image region perceived in accordance with WLpeephole as its input (i.e., the first layer of the deep network), and at its last layer outputs a scalar value indicating the category of the corresponding region, namely waterline or non-waterline. Moreover, apart from its input and output, the overall architecture of WLdetectNet is a linear stack of five repeatable structural modules, which totally consists of 72 convolutional layers. Because of following the common architectural principles with modern deep CNNs, WLdetectNet is easy to be constructed and trained. 

\subsection{The auxiliary network WLgenerateNet}
To guarantee the stability of our waterline detection approach under varied inland water environments (i.e., its robustness), in the proposed paradigm DeepWL, we intentionally arrange another deep network named WLgenerateNet as an auxiliary network to assist the main network WLdetectNet in improving its generalization ability. Moreover, the accuracy and efficiency of WLdetectNet continue to be maintained. Specifically, we construct the WLgenerateNet by following the design principles of GANs, and then utilize it to build on demand a large amount of waterline samples relevant to various scenarios for training WLdetectNet, thus generalizing the representational capability of the WLdetectNet and enabling the main network of our paradigm to be effectively applicable for waterline detection in diverse scenarios. That is motivated by a fact in machine learning: more data samples help to improve the generalization ability of a model (e.g., CNNs) and mitigate its problem of overfitting, thus improving the robustness of the model. However, it is actually not easy to collect such a large amount of labeled data on various waterlines. Therefore, in our waterline detection approach, we ingeniously draw lessons from the spirit of GANs that they can enable the automatic generation of desired data. 

Similar to classic GANs, the WLgenerateNet consists of two convolutional neural networks contesting with each other in a zero-sum game framework, where the two adversarial networks are a generator \emph{G(z)} for generating waterline samples and a discriminator \emph{D(x)} for discriminating waterline samples, respectively.

In the WLgenerateNet, we utilize a 100-dimensional random noise $z$ as its input, then convert $z$ into a 64x64 pixel image $x$ by generator \emph{G(z)}. Meanwhile, discriminator \emph{D(x)} is applied to determine whether the currently generated image $x$ belongs to waterline. Just in the case that the result of \emph{D(x)} is true, the WLgenerateNet outputs generated images. Finally, through an iterative process of \emph{G(z)} and \emph{D(x)} contesting with each other, we can gain our desired labeled data on waterline. 

\subsection{Training methods of two deep networks}
As illustrated above, the proposed paradigm DeepWL comprises two specially designed deep neural networks, i.e., WLdetectNet and WLgenerateNet. Here, we focus on their training methods, in which the WLdetectNet is trained in a supervised learning fashion while the WLgenerateNet is trained in an unsupervised learning fashion. 

\emph{1) Training WLdetectNet by supervised learning}:
The WLdetectNet acts as the mainstay of DeepWL. According to its design schemes described previously, WLdetectNet aims to capture discriminative information relevant to waterline segments for final waterline detection. Thereby, in order to guarantee its accuracy and generalization ability, a large amount of training data is required, except for those significant designs regarding its architecture. However, no public dataset on waterline is available at present. Moreover, as mentioned before, it is also very difficult to collect such a large dataset, due to the labor and economic costs. To effectively carry out the training of WLdetectNet, we opt to build our own dataset on waterline segments in a simple and economical manner, which involves the following two processes. 

 \emph{ Manually gathering original data satisfying the structural layout of WLpeephole}.
We first gather 2,000 image patches containing diverse waterline segments by manually cropping from surveillance videos associated with inland waterline, then resize them to be consistent with the structural layout of WLpeephole, especially compelling their waterline segments to display only in a smaller scope same with the recognizing field of WLpeephole. In practice, these gathered image patches come from varied scenarios including dissimilar weather conditions and different illumination conditions, so that the diversity of these samples is helpful for enhancing the perceptive ability of WLpeephole to detect various waterline segments, thus improving the generalization ability of WLdetectNet. 

\emph{Automatically generating artificial data on waterline segments for data augmentation}.
Due to labor and economical costs, it has been proved to not be an easy thing that hunting for plentiful labeled data on waterline segments. Thus, aside from manually gathering more such samples that are representative of distinct waterline segments, we also attempt to augment the labeled data we already have by means of WLgenerateNet. Specifically, we make advantage of WLgenerateNet to automatically generate more artificial data on waterline segments (almost 8,000 image patches at present) from existing manual dataset (i.e., 2,000 image patches). In fact, our approach to data augmentation by GANs on images is great for combating overfitting that is one of the primary problems with machine learning models in general, since we can further enlarge these data on demand.

Through the above two processes of manually gathering and automatically generating, around 10,000 image patches on waterline segments have constituted the positive samples of our training set. Furthermore, the training set also contains about 12,000 negative samples that are freely cropped from various non-waterline images. Importantly, the training set is a scalable and generalizable dataset, since it can further generalize and augment its sample data according to the needs of practical applications by the two processes mentioned above. 

Then, based on the built dataset, we conduct the training for WLdetectNet by minimizing an energy function, which can be formally expressed as: 

\begin{footnotesize}
\begin{equation}
\mathrm{E}(\theta)=-\sum_{\mathrm{i}=1}^{N}\left[y_{i} \ln f_{\theta}\left(x^{i}\right)+\left(1-y_{i}\right) \ln \left(1-f_{\theta}\left(x^{i}\right)\right)\right]
\end{equation}
\end{footnotesize}

\noindent where \emph{N} is the number of training samples, $\theta$  refers to all parameters of the deep learning model WLdetectNet, $x^{i}$ denotes the $i^{th}$ training sample, $f_{\theta}\left(x^{i}\right)$  denotes the output of WLdetectNet over  $x^{i}$ whose architecture is described in Table \ref{tab1}, and  $y^{i}$ represents the ground-truth label of sample  $x^{i}$ where scalar-valued 1 for waterline and 0 for non-waterline. Our optimization goal for this energy function is to chase the sweet spot where the cross entropy loss of WLdetectNet is low when its parameters are tuned by stochastic gradient descent (SGD) algorithm with a batch size of 60. Moreover, to avoid gradient explosions, our training procedure for WLdetectNet is divided into two stages: we first employ the samples gathered manually and 3,000 negative cases to train WLdetectNet for 30 epochs, then apply the data generated by WLgenerateNet and 9,000 negative cases to fine-tune WLdetectNet for 50 epochs. Finally, a binary classification deep network to detect waterline segments based on WLpeephole is obtained. 

\emph{2) Training WLgenerateNet by unsupervised learning}:
As an auxiliary facility in DeepWL, WLgenerateNet aims to support the generalization ability of another deep learning model WLdetectNet by rendering more training data as much as possible for WLdetectNet. Similar to other classic GANs, its learning objective corresponds to a minmax two-player game, which is formulated as:

\begin{footnotesize}
\begin{equation}
\min _{G} \max _{D} \mathrm{L}(G, D)=\mathbb{E}_{x \sim p_{\text {datd}}(x)}[\log D(x)]+\mathbb{E}_{z \sim p_{z}(z)}[\log (1-D(G(z)))]
\end{equation}
\end{footnotesize}

\noindent where the generator \emph{G(z)} is responsible for learning to map data $z$ from the noise distribution $P_{z}(z)$ to the distribution $p_{\operatorname{dat} a}(x)$ over data $x$, while the discriminator \emph{D(x)} answers for estimating the probability of a sample from the data distribution $p_{\operatorname{data}}(x)$ rather from \emph{G(z)}. 

\subsection{Our waterline detection algorithm based on DeepWL}

As stated before, WLdetectNet performs as the mainstay of the paradigm DeepWL for our vision-based waterline detection, whereas its receptive field is constrained to a same region as WLpeephole whose size is fixed in practical applications. Thereby, our proposed paradigm DeepWL is more applicable for distinguishing segments relevant to waterline, i.e., determining if there is a waterline segment in the detected image region. 

Given that the size of an image captured from USVs may be arbitrary, we present a waterline detection algorithm based on DeepWL, which pursues waterline image-map estimation from single video stream captured on board. The algorithm is summarized in Algorithm 1, which results in a corresponding waterline estimation image-map (see Fig.\ref{fig1}). In the algorithm, we formulate the task on waterline image-map estimation as a sequence of repetitive subtasks on distinguishing waterline segment in every handled video frame, wherein each subtask is conducted by taking advantage of DeepWL that implicitly consists of two stages: estimating potential waterline segment via WLpeephole and marking associated waterline segment via a specific strategy. Indeed, a waterline usually can be deemed to the combination of a spectrum of line segments.

\begin{algorithm}
\caption{Waterline Detection Algorithm} 
\label{alg1}
\begin{algorithmic}
\REQUIRE  
\STATE
Single video stream $X=\left\{x_{t}\right\}_{t=1 : k}$, sampling rate $f$ , scale of WLpeephole $r$ , stride of WLpeephole moving in every image $h$, WLdetectNet and its learned parameters $C_{W}$.
\ENSURE 
\STATE
 A sequence of waterline estimation image-maps $Y=\left\{y_{i}\right\}_{i=1: k / f} $ \\
\end{algorithmic}
\textbf{Procedure:}\\ \hangindent 
1.5em 	1: Sample $X=\left\{x_{t}\right\}_{t=1 : k}$ online according to sampling rate $f$, which can eventually derive a corresponding sequence of images  $Z=\left\{z_{i}\right\}_{i=1 : k / f}$.\\
	2: Take the currently derived frame $z_{i}$ by sampling as an image to be detected. \\
	3: Initialize the placement of WLpeephole within $z_{i}$ to be at the upper left corner of $z_{i}$.\\
	4: Fetch the image region (denoted as $b$) corresponding to WLpeephole as the current receptive field of WLdetectNet $C_{W}$.\\
	5: Apply WLdetectNet $C_{W}$ to estimate if there is a waterline segment in $b$, i.e., derive the value of $C_{W}(b)$.\\
	6: If the label of $C_{W}(b)$ corresponds to waterline, mark $b$ according to a strategy: mark the central pixel of $b$, else not. \\
	7: Move WLpeephole (vertically or horizontally) to a new placement within $z_{i}$, according to stride $h$.\\
	8: Iterate steps 4 to 7 until WLpeephole moving to the lower right corner of $z_{i}$.\\
	9: Connect all marked pixels within $z_{i}$ as an estimated waterline, then output $z_{i}$ to be current waterline estimation image-map $y_{i}$.\\
	10: Iterate steps 2 to 9 until $i=k / f$.
\end{algorithm}

Thus, the performance of the algorithm depends to a large extent on the main network of our paradigm DeepWL, whose effectiveness and efficiency are put forward in this paper by the aforementioned two significant schemes relevant to WLdetectNet with the assistance of WLgenerateNet. Furthermore, in order to accelerate waterline image-map estimation from single video stream, we do not resort to handling every frame of single video stream in the algorithm. Meanwhile, in case of needing to present more fine-grained marking effect of waterline in every estimation image-map, we can also opt a more considerate marking strategy to the algorithm. Notwithstanding in Algorithm 1 we employ a relatively simple marking strategy to rapidly approximate potential waterlines, it is actually enough for demands within inland waters. 

\section{Experiments and evaluation}
\label{sec3}
To demonstrate the effectiveness and superiority of the proposed deep waterline detection approach, related experimental assessment and results are presented. 

\subsection{Experimental settings}
The proposed waterline detection approach (Algorithm 1) has been deployed in our own USV customized to patrol within inland water, which is equipped with a visual perception subsystem for assisting navigation. All experimental data (optical images) were acquired by a general digital camera mounted in its visual perception subsystem with auto focus and exposure mode, when it was travelling in the East Lake, one of the largest urban lakes in China. Importantly, these obtained data, whose resolutions are 1080x1440, came from varied waterfront scenarios under different weather and illumination conditions, such as sunset with weak illumination, sunny weather with strong illumination, and foggy weather.

\subsection{Evaluation metrics}
To enable the quantitative assessment of performances on different waterline detection algorithms (or systems), relevant evaluation metrics are indispensable. Since there is no specialized metric to evaluate vision-based waterline detection, we establish several necessary statistical indicators to measure the performances of concern to us, by referring to the evaluation methods for classification models.

 \emph{Effectiveness.}
To verify the performance of a waterline detection algorithm, its effectiveness in a waterline estimation image-map needs to be proved first. We adopt \emph{precision-recall} metrics to characterize the detection effectiveness, which calculate how close the estimated results compare with the ground truth. Formally, \emph{precision} and \emph{recall} are defined as follows: 

\begin{footnotesize}
\begin{equation}
precision=\frac{\text {Card.of }\left\{e_{i} | \forall_{i, j} d i s t\left(e_{i}, a_{j}\right) \leq \lambda, \text {and } i, j \in \mathbb{N}\right\}}{\text { Card.of a finite set }\left\{e_{i} | i \in \mathbb{N} \right\}}
\end{equation}
\end{footnotesize}

\begin{footnotesize}
\begin{equation}
recall=\frac{\text {Card.of }\left\{e_{i} | \forall_{i, j} d i s t\left(e_{i}, a_{j}\right) \leq \lambda, \text {and } i, j \in \mathbb{N}\right\}}{\text { Card.of a finite set }\left\{g_{i} | i \in \mathbb{N}\right\}}
\end{equation}
\end{footnotesize}

\noindent where card. refers to the cardinal of a finite set, $e_{i}$ denotes each pixel that lies within an estimated waterline, $a_{j}$ denotes each anchor marked manually in original image, all of which are connected to be a ground-truth waterline, and eventually the ground truth by hand results in a finite set consisting of a sequence of relevant pixels $g_{i}$ in original image. Besides, \emph{dist}($e_{i}$, $a_{j}$) refers to the distance of image coordinate between $e_{i}$ and $a_{j}$, and $\lambda$ represents a threshold on visual distance, which is specifically set according to practical scenarios. 

\emph{Robustness.}
Environmental variations, such as weather, illumination and water condition, often interfere with the effect of waterline detection. For instance, in some scenarios, waterline detection obtains ideal recall, whereas its precision demonstrates the opposite. The reason is that many pixels irrelevant to waterline have been mistakenly detected as the ground truth. Thus, we need to test the robustness of waterline detection against environmental noises so that the capacity of resisting environmental disturbances to a certain waterline detection approach can be better analyzed. To this end, we define \emph{FP}-irrelevance metrics to quantify the robustness of waterline detection against environmental noises in an estimated image-map, wherein \emph{FP} counts the number of all false positives, i.e., how many irrelevant pixels caused by noises are selected in an image-map, and irrelevance measures the overall deviation trend of pixel-level distances between those irrelevant pixels and the ground truth, which actually characterizes the statistical distribution on the distances of irrelevant pixels related to the ground truth in an estimated image-map. Formally, \emph{FP} and \emph{irrelevance} are defined as follows: 

\begin{footnotesize}
\begin{equation}
F P=\operatorname{Card.of}\left\{e_{i} | \min \left\{\forall_{j} d i s t\left(e_{i}, a_{j}\right)\right\}>\lambda, \text {and }  i, j \in \mathbb{N}\right\}
\end{equation}
\end{footnotesize}

\begin{footnotesize}
\begin{equation}
irrelevance=S K \text { of }\left\{d_{i} | d_{i}=\min \left\{\forall_{j} \operatorname{dist}\left(e_{i}, a_{j}\right)\right\}>\lambda, \text {and } i, j \in \mathbb{N}\right\}
\end{equation}
\end{footnotesize}

\noindent where \emph{min\{*\}}  denotes the minimum of all elements in a finite set \emph{\{*\}}, \emph{SK} refers to the asymmetry coefficient of the skewness distribution on the pixel-level distance between the wrongly estimated waterline and the ground truth, and the set $\left\{e_{i}\right\}$ in Eq. (5) actually represents a finite set consisting of all irrelevant pixels in an estimated image-map. Besides, $e_{i}$, $a_{i}$, $\lambda$, and \emph{dist}$\left(e_{i}, a_{j}\right)$  are similar to the ones in Equations (3) and (4). 

As far as an evaluated waterline detection approach is concerned, in the case of same \emph{FP}, if the distance distribution presents positive skewness and higher irrelevance is obtained, we consider those wrongly estimated pixels to be more convergent to ground truth, and further its robustness is deemed to be better. In other words, in this case, the evaluated approach enables the impact from environmental disturbances on waterline detection effect to be shrunk as far as possible into the area around ground truth, where its estimation error gets smaller. Correspondingly, its capability to withstand environmental noises manifests more robust. 

\emph{Stability.}
For continuous waterline detection based on video, we often need to inspect the impact of environmental variations on a sequence of estimated image-maps when facing the same visual scenario. Thereby, a related metric called \emph{stability} is defined to quantify the stability of an evaluated approach under changeable environments. Specifically, the \emph{stability} involves measuring the stability over four different metrics (\emph{precision, recall, FP} and \emph{irrelevance, respectively}) on multiple estimated image-maps, when a waterline detection approach is evaluated for a specific scenario against different environmental noises. Formally, \emph{stability} over a metric $p$ is defined as follows: 

\begin{footnotesize}
\begin{equation}
stability(p)=\frac{\operatorname{mean}(p)-m e d i u m(p)}{\sigma(p)}
\end{equation}
\end{footnotesize}

\noindent where $p$ denotes the metric \emph{precision}, \emph{recall}, \emph{FP} or \emph{irrelevance}, $mean(p)$ denotes the mean of a specific metric \emph{p} over all assessed samples (i.e., estimated image-maps for the same scenario), $medium(p)$ and $\sigma(p)$ refer to the medium and standard deviation of those samples relevant to $p$, respectively. 

In essence, \emph{stability} characterizes four distribution conditions on their corresponding metrics by sampling diverse estimated image-maps that represent respective results from those facing same visual scenario with varied environmental noises. Given a metric $p$, if $stability(p)$ tends to be zero, the results about the specified metric over all samples are more convergent to normal distribution, which means that evaluated approach has more stability on this metric against environmental variations. On the contrary, the results with respect to the metric are prone to be fragile for environmental noises. 

\subsection{Results and analysis on our deep waterline detection approach}

\begin{figure}[htbp]
\centerline{\includegraphics[width=7.5cm]{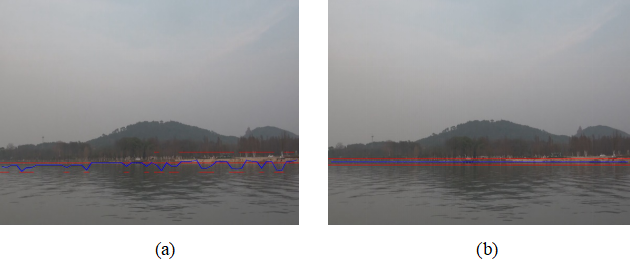}}
\caption{Visual comparisons on detection results by performing Algorithm 1 on a single image with two different scales of WLpeephole (here, blue line shows the estimated waterline, and red line acts as the subline for marking manually the ground truth): (a) \emph{r}=30 and \emph{s}=10, (b) \emph{r}=60 and \emph{s}=30}
\label{fig4}
\end{figure}

The scale of WLpeephole includes the size of \emph{observing field} (denoted as \emph{r}) and the size of \emph{recognizing field} (denoted as \emph{s}), which  are two significant impact factors of our detection algorithm (Algorithm \ref{alg1}). To validate the effectiveness of the proposed approach to waterline detection, we conduct a group of experiments from the perspective of investigating the two impact factors on the resulting accuracy. Specifically, we carry out our algorithm repeatedly on the same optical image (presented in Fig. \ref{fig4}) in the case of 10 different (\emph{r}, \emph{s}) pairs. Notably, since both higher precision and higher recall are usually expected for practical waterline detection tasks, here we employ \emph{F1-score} to evaluate the effectiveness of our approach on a single optical image. In practice, \emph{F1-score} depends on \emph{precision-recall} metrics, which is generally formulated as below: 

\begin{footnotesize}
\begin{equation}
F 1=\frac{2 \times p r e c i s i o n \times r e c a l l}{p r e c i s i o n+r e c a l l}
\end{equation}
\end{footnotesize}

Then, ten relevant \emph{F1-scores} are calculated, as shown in Table \ref{tab3}. Among them, Fig. \ref{fig4}(a-b) illustrates the visual result in the case of (48, 24) and (60, 30) for (\emph{r}, \emph{s}) pair, respectively. 

\begin{table}[htbp]
\caption{ Quantitative comparisons on \emph{F1-score} by setting ten different scales of WLpeephole respectively (here, \emph{H} denotes the height of an image, $\lambda$=10 pixels)}
\begin{center}
\begin{tabular}{p{17pt}<{\centering}|p{10pt}<{\centering}|p{10pt}<{\centering}|p{13pt}<{\centering}|p{10pt}<{\centering}|p{10pt}<{\centering}|p{13pt}<{\centering}|p{10pt}<{\centering}|p{10pt}<{\centering}|p{13pt}<{\centering}}
\hline
\multirow{2}{*}{\textbf{scales} }
& \multicolumn{3}{c|}{\textbf{\emph{r=H/36}}} & \multicolumn{3}{c|}{\textbf{\emph{r=H/22.5}}} & \multicolumn{3}{c}{\textbf{\emph{r=H/18}}}   \\
\cline{2-10}
& \textbf{\emph{r}}&\textbf{\emph{s}}&\textbf{\emph{F1}}&\textbf{\emph{r}}&\textbf{\emph{s}}&\textbf{\emph{F1}}&\textbf{\emph{r}}&\textbf{\emph{s}}&\textbf{\emph{F1}} \\
\hline
s=r/3&30&10&\textbf{0.685}&48&16&0.841&60&20&0.915 \\
\hline
s=r/2&30&15&0.712&48&24&0.865&60&30&\textbf{0.943} \\
\hline
s=2r/3&30&20&0.706&48&32&0.858&60&40&0.929 \\
\hline
\end{tabular}
\label{tab3}
\end{center}
\end{table}

From Table \ref{tab3}, we observe that our waterline detection algorithm is effective in practical inland water scenario, even though the scale of WLpeephole impacts on its resulting accuracy more or less. Among of the ten displayed \emph{F1-scores}, the one (i.e., 0.943) is highest when (\emph{r}, \emph{s}) pair is set to (60, 30), which actually represents the best detection effect that has been attained in this group of experiments, just as shown in Fig.\ref{fig4}(b). The reason is that, in this case, more contextual information relevant to waterline and more sufficient information about waterline itself have been fed into our waterline discriminator WLdetectNet, which benefit from our having chosen bigger and more appropriate receptive field as far as possible by the current (\emph{r}, \emph{s}) pair. Instead, Fig.\ref{fig4}(a) shows the worst visual result that corresponds to the case of (30, 10) in Table \ref{tab3}, in which \emph{F1-score} presents the lowest (i.e., 0.685). However, too big receptive field also decays the detection effect. For example, when (\emph{r}, \emph{s}) pair is set to (90, 45), its \emph{F1-score} gets just 0.835. It is because that our marking strategy to approximate potential waterlines in Algorithm 1 is simplistic, so that fine-grained marking effect in an estimated image-map is difficult to achieve in the case of setting such big scale of WLpeephole. Subsequently, the accuracy of detection suffers more frustration. As a result, we suggest that the (\emph{r}, \emph{s}) pair in Algorithm 1 can be empirically set to (60, 30), which is usually a good choice for practical applications based on our waterline detection algorithm, especially for detecting a 1080x1440 image.

\subsection{Comparison to other alternative approaches}

To verify the superiority of the proposed waterline detection approach, we carry out experimental comparison between relevant alternative approach and ours. 

Current vision-based waterline detection primarily resorts to non-deep-learning methods. Specifically, they generally apply a non-deep-learning paradigm to focus on waterline-relevant feature representation or final discriminative strategy (algorithm). Among them, edge detection is such a classic method that has been extensively applied in applications based on waterline detection. Thus, in this subsection, we compare the representative method with ours on their robustness and stability in the presence of environmental noises. 

\noindent \emph{a) Visual comparison}

As very common environmental noises to waterline detection within inland water, four environmental interference factors are paid attention to in our experiments, which are linear objects, water ripples, shadow and fog. Here, we are primarily concerned about their impacts on the detection results. 

\begin{figure}
    \begin{subfigure}[b]{0.115\textwidth}
        \includegraphics[width=\textwidth]{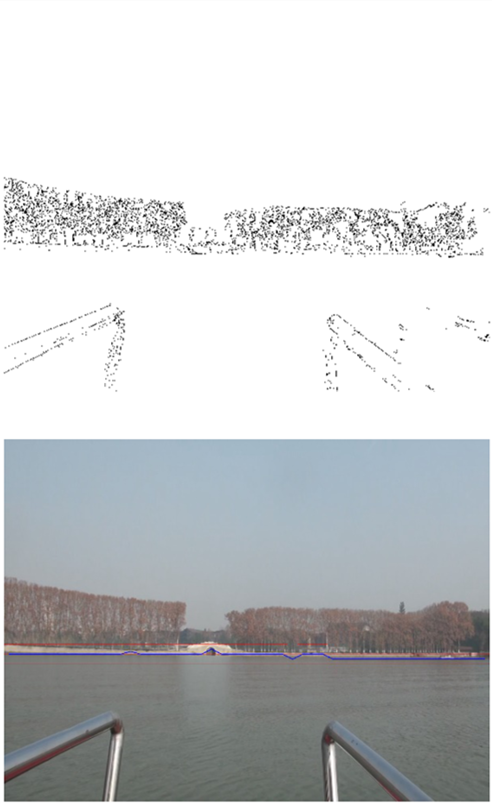}
        \caption{}
        \label{fig:visual_smap_o}
    \end{subfigure}
    \begin{subfigure}[b]{0.115\textwidth}
        \includegraphics[width=\textwidth]{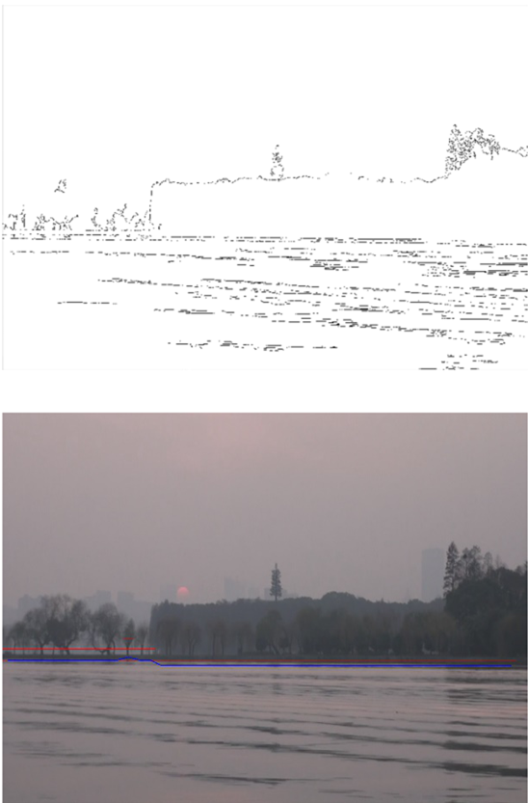}
        \caption{}
        \label{fig:visual_smap_c}
    \end{subfigure}
    \begin{subfigure}[b]{0.115\textwidth}
        \includegraphics[width=\textwidth]{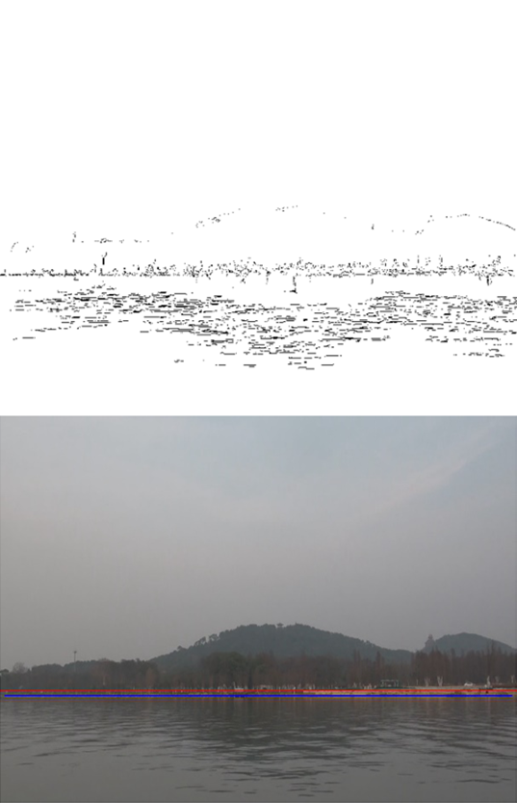}
        \caption{}
        \label{fig:visual_smap_gt}
    \end{subfigure}
	\begin{subfigure}[b]{0.115\textwidth}
        \includegraphics[width=\textwidth]{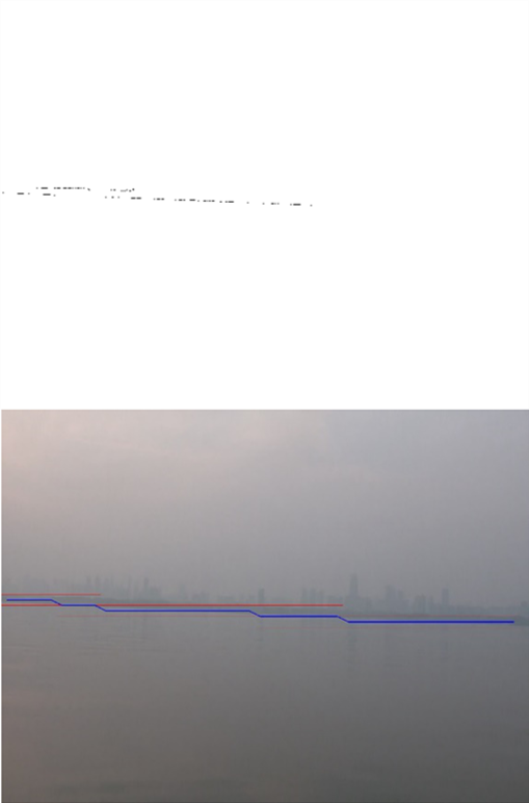}
        \caption{}
        \label{fig:visual_smap_gt}
    \end{subfigure}
    \caption{Visual comparison of results between edge detection method (upper) and ours (bottom) with different environmental interference variables: (a) linear objects, e.g., rails of the USV, (b) water ripples, (c) shadow, (d) fog. (In our results, blue line shows the estimated waterline, and red line acts as the subline for marking manually the ground truth)}
\label{fig5}
\end{figure}

Figure \ref{fig5} shows the visual results by Canny edge detector and ours against our concerned environmental noises on waterline detection. In the first row of Fig.\ref{fig5}, all elements in black depict the estimated waterlines by Canny edge detector. And, the second row of Fig.\ref{fig5} presents our results, in which blue line indicates the estimated waterlines by our approach. 

From Fig.\ref{fig5}, it is observed that our results are obviously better than the compared approach in terms of handling environmental noises. For example, at the top of Fig.\ref{fig5}(a-c), rails of our USV, parts of water ripples and shadow are wrongly detected as waterlines, and at the top of Fig.\ref{fig5}(d), real waterline are not completely detected due to low visibility. In contrast, our estimated waterlines at the bottom of Fig.\ref{fig5}(a-d) are basically concentrated in the vicinity of ground truth. Therefore, in terms of resisting environmental disturbance, our approach is intuitively superior to the alternative approach. 

\noindent \emph{b) Quantitative assessment on robustness}

Then, according to Eq.(3-6), we have calculated the \emph{precision-recall} metrics and \emph{FP-irrelevance} metrics respectively corresponding to the visual results presented in Fig.\ref{fig5}. The quantitative comparisons on these evaluated metrics are shown in Table \ref{tab5}. 

\begin{table}[htbp]
\caption{Quantitative comparisons of assessed metrics corresponding to the results by edge detection method (left of /) and ours (right of /) under different noises}
\begin{center}
\begin{tabular}{p{37pt}<{\centering}|p{41pt}<{\centering}|p{41pt}<{\centering}|p{41pt}<{\centering}|p{33pt}<{\centering}}
\hline
\textbf{Metrics}&\multicolumn{4}{c}{\textbf{Environmental interference factors}} \\
\cline{2-5} 
{$\lambda=10$} & \emph{rail} & \emph{ripple} & \emph{shadow} & \emph{fog} \\
\hline
\emph{precision}(\%) &	11.5/96.5 &	29.2/95.4	& 21.2/96.1  &	93.3/90.6 \\

\emph{recall}(\%)  &	97.6/98.2 &	95.2/97.1 &	83.7/98.6 &	41.9/84.5 \\

\emph{FP(pixels)} &	 $\sim$54K/40 &	$\sim${}32K/36 &	$\sim${}33K/38 &	16/43 \\

\emph{irrelevance} &	-1.119/0.092 &	-1.413/0.068 &	-1.216/0.073 &	0.059/0.081 \\
\hline
\end{tabular}
\label{tab5}
\end{center}
\end{table}

Usually, for a robust waterline detection approach, both high precision and high recall are desired in any water environments. From Table \ref{tab5}, we see that the edge detection method (Canny edge detector) attains same desirable recalls as ours when rail, ripple or shadow is emerging, whereas its corresponding precisions are much lesser than ours. Besides, under foggy weather conditions, despite both of the two compared methods obtain high precision, the recall of edge detection method is only half of ours. Obviously, for the same scenarios, the effectiveness of edge detection method is more sensitive to environmental noises than ours. The reason is that a large number of pixels irrelevant to waterline are also selected by edge detection method, while correct pixels are annotated in an estimated image-map. For instance, there are 54,182 false positives (\emph{FP}) also marked as black pixels in the upper image of Fig.\ref{fig5}(a). Moreover, although the number of irrelevant pixels is rather small (only 16 false positives) owing to low illumination caused by fog, many true positives are still missed in the result by edge detection method that induces unsatisfactory recall (just 41.9\%). Then, the measured \emph{irrelevances} shown in Table \ref{tab5} indicate that most of \emph{irrelevance} metrics on edge detection method are negative, which means those irrelevant pixels (false positives) selected by this method scatter around real waterline. On the contrary, our \emph{irrelevance} metrics are positive, implying that our false positives as a whole are more approximate to the ground truth. 

Actually, in practical waterline detection tasks, edge detection method, as well as many other alternative approaches such as waterline detection based on image segmentation, generally employs necessary image preprocessing (e.g., image denoising) to eliminate those irrelevant pixels induced by environmental noises and guarantee high precision and high recall at the same time. However, these approaches relying on image preprocessing influence the efficiency of waterline detection tasks more or less due to extra computational costs. Instead, our approach can straightforwardly distinguish candidate waterline segments from raw images, since the proposed approach adopts an end-to-end paradigm based on deep learning, in which there is no preprocessing procedure. Therefore, as far as a single estimated image-map is concerned, our approach has better capacity of resisting environmental disturbance in the absence of image preprocessing, which is also demonstrated by the visual comparisons presented in Fig.\ref{fig5}. 

\noindent \emph{c) Quantitative assessment on stability}

To evaluate the stability of waterline detection approach on video data, especially at the moments when environmental factors (e.g., weather or illumination conditions) emerge variations in a visual surveillance scenario, we further conduct related experimental comparisons between edge detection approach and ours. Here, we take foggy condition bringing about illumination variation as an example, and test its impacts on the stabilities of the two approaches during the procedure that time-sequence images are successively dealt with. Correspondingly, we sample 150 image frames from eight hours of video, which cover varied foggy conditions in the same monitoring scenario. Then, based on the \emph{precision-recall} metrics and \emph{FP-irrelevance} metrics associated with each of these samples that are achieved respectively in terms of Eq. (3-6), we calculate the \emph{stability} over the previous four metrics according to Eq. (7), as shown in Table \ref{tab6}. 

\begin{table}[htbp]
\caption{Quantitative comparisons on \emph{stability} for edge detection method (with image preprocessing) and ours under varied foggy conditions}
\begin{center}
\begin{tabular}{c|c|c}
\hline
\textbf{\emph{Stability}} over&\textbf{Canny edge detector}&\textbf{Ours} \\
\hline
\emph{precision} &	-1.767 &	-1.153 \\

\emph{recall} &	-3.198 &	0.991 \\

\emph{FP} &	1.052 &	-1.124 \\

\emph{irrelevance} &	-0.313 & 	0.196 \\
\hline
\end{tabular}
\label{tab6}
\end{center}
\end{table}

In the experiment about stability assessment, to achieve more impartial effect, we practically employ classic Canny edge detector with necessary image preprocessing as an evaluated edge detection method to compare with ours. From Table \ref{tab6}, we can see that the measurements of our approach regarding \emph{stability} over our concerned metrics are closer to zero than Canny edge detector with image preprocessing, except for the metric \emph{FP} due to more irrelevant pixels caused by our approach. It indicates that our measuring results over these samples with respect to most of our concerned metrics, e.g., \emph{precision}, \emph{recall} and \emph{irrelevance}, are more convergent to normal distribution. Therefore, our approach has more stability on the corresponding metrics against environmental variations. 

\section{Conclusion}
\label{sec4}
To respond to the challenge from highly dynamic inland water environments, maritime applications require an onboard vision-based waterline detection algorithm with more robustness and more stability to accomplish specific missions. In this paper, we proposed a novel visual detection approach to identifying inland waterlines with general digital camera by the use of deep learning techniques, which aimed to guarantee the effectiveness of waterline detection within variable inland water environments. Meanwhile, to evaluate our concerned performances, we defined quantitative metrics and conducted empirical investigations. Experimental results in real-life scenarios demonstrated that our approach performed more favorably than the compared approach, and achieved better robustness and stability in the presence of visual noises in dynamic inland waters. In addition, due to the generality of our proposed approach, it is also suitable for the waterline detection tasks of other water areas, such as coastal waters.

\bibliographystyle{IEEEtran}
\bibliography{IEEEabrv,referencesTest}

\end{document}